%% file: Template_ISBI_latex.tex
\documentclass{article}
\usepackage{spconf,amsmath,graphicx}
\usepackage[T1]{fontenc}
\usepackage{changes}
\usepackage[noadjust]{cite}
\usepackage{cite}
\usepackage[shortlabels]{enumitem}
\usepackage{tikz,pgf} 
\usetikzlibrary{arrows,automata} 
\usepackage{verbatim}
\usetikzlibrary{positioning,shapes.geometric}
\usepackage{changes}
\usepackage{amsmath}
\usepackage{algpseudocode}
\usepackage{tabularx}
\usepackage{array}
\usepackage{float}
\usepackage{amsmath}
\usepackage{amssymb}
\usepackage{algorithm}
\usepackage{mathtools,xparse}

\usepackage{amssymb}
\usepackage{gensymb}
\usepackage{array}
\usepackage{tabularx}
\usepackage{enumitem}
\usepackage{lipsum}
\usepackage{color}
\usepackage{graphicx}
\usepackage[caption=false]{subfig}
\usepackage{amsmath}
\usepackage[all,poly]{xy}
\usepackage{algpseudocode}
\usepackage{array}
\usepackage[]{siunitx}
\usepackage{textcomp}
\usepackage{nicematrix}
\usepackage[english]{babel}
\usepackage{blindtext}
\usepackage{adjustbox}
\usepackage{cite}
\input com_notations.tex

\input notations_yoann.tex

\title{Direct Low-Field MRI Super-Resolution using undersampled k-Space}
\name{Daniel Tweneboah Anyimadu$^{(1)}$, Mohammed M. Abdelsamea$^{(1)}$, and Ahmed Karam~Eldaly$^{(1)(2)}$}
\address{$^{(1)}$Department of Computer Science, University of Exeter, Exeter, United Kingdom\\ $^{(2)}$UCL Hawkes Institute, University College London, London, United Kingdom}

\begin{document}
%
\maketitle
\begin{abstract}
\vspace{-0.1cm}
Low-field magnetic resonance imaging (MRI) provides affordable access to diagnostic imaging but suffers from prolonged acquisition and limited image quality. Accelerated imaging can be achieved with \emph{k}-space undersampling, while super-resolution (SR) and image quality transfer (IQT) methods typically rely on spatial-domain post-processing. In this work, we propose a novel framework for reconstructing high-field MR like images directly from undersampled low-field \emph{k}-space. Our approach employs a \emph{k}-space dual channel U-Net that processes the real and imaginary components of undersampled \emph{k}-space to restore missing frequency content. Experiments on low-field brain MRI demonstrate that our \emph{k}-space-driven image enhancement consistently outperforms the counterpart spatial-domain method. Furthermore, reconstructions from undersampled \emph{k}-space achieve image quality comparable to full \emph{k}-space acquisitions. To the best of our knowledge, this is the first work that investigates low-field MRI SR/IQT directly from undersampled \emph{k}-space.
\end{abstract}
\vspace{-0.25cm}
\section{Introduction}
\label{sec:intro}
\vspace{-0.25cm}
Magnetic Resonance Imaging (MRI) underpins a wide range of diagnostic and research workflows by offering excellent soft-tissue contrast and non-invasive imaging. While high-field (HF) scanners ($\ge 1.5$T) deliver superior signal- and contrast-to-noise ratios, their cost and infrastructure demands restrict global access. Low-field MRI (LF-MRI; $<\!1$T) is a practical alternative that is gaining traction for point-of-care and resource-limited settings \cite{murali2024bringing}. However, LF-MRI faces two persistent challenges: longer scan times and intrinsically lower signal to noise ratio (SNR) which degrades image quality and clinical reliability \cite{arnold2023low}.
 
It is crucial to recall that MRI data are acquired in the frequency domain (\emph{k}-space). Fully sampling \emph{k}-space (e.g., $\bfy=\bfF\bfx$, where $\bfy$ is the vectorised \emph{k}-space measurements, $\bfx$ is the corresponding spatial domain image, and $\bfF$ is the Fourier transform) is time-consuming; practical protocols adopt undersampling with a sampling pattern $\bfOmega$ (conceptually $\bfy=\bfF_{\bfOmega}\bfx$), which accelerates acquisition but yields an ill-posed inverse problem that further degrades image quality \cite{jhamb2015review}. Compressed sensing and deep learning have, however, advanced accelerated reconstruction using undersampling \cite{eldaly2024bayesian,Eldaly2026ICASSP,safari2025advancing}. 

On the other hand, super-resolution (SR) and image quality transfer (IQT) learn mappings between high-quality reference scans and the counterpart low-quality ones to recover missing information in LF-MRI scans \cite{alexander2014image, alexander2017image, lin2023low, kim20233d, eldaly2024alternative, Tien2026ISBI}. While SR/IQT has shown promise for enhancing LF-MRI, existing approaches operate exclusively in the spatial domain as a postprocessing step following image reconstruction and ignore the raw \emph{k}-space data \cite{alexander2014image,alexander2017image,lin2023low,kim20233d,eldaly2024alternative}. This separation between reconstruction and quality enhancement potentially limits performance, as valuable frequency-domain information is discarded before SR/IQT is applied. In this work, we propose to address this gap by performing SR/IQT directly in the \emph{k}-space domain, thereby integrating image reconstruction and SR/IQT into a unified framework. Our method enables the recovery of missing frequency content and reconstructs HF-MR like images. Thus, the main contributions of this work are as follows. (1) We propose a novel framework that reconstructs HF-like MR images from undersampled LF-MR \emph{k}-sapce measurements. This is the first attempt in the literature that combines LF-MRI reconstruction and SR/IQT in a unified framework. (2) We propose a dual channel U-Net architecture that jointly processes real and imaginary channels of \emph{k}-space, enabling accurate restoration of missing frequency information. This leverages the frequency domain's phase and magnitude consistency, which is typically lost in traditional spatial-domain LF-MRI SR and IQT. (3) We demonstrate that our \emph{k}-space-driven SR/IQT approach achieves superior performance over the counterpart spatial-domain method, and yields reconstructions from undersampled data that are comparable in quality to full \emph{k}-space acquisitions.
\vspace{-0.3cm}
\section{Method}
\label{sec:method}
\begin{figure}
\centering
\includegraphics[width=0.995\columnwidth]{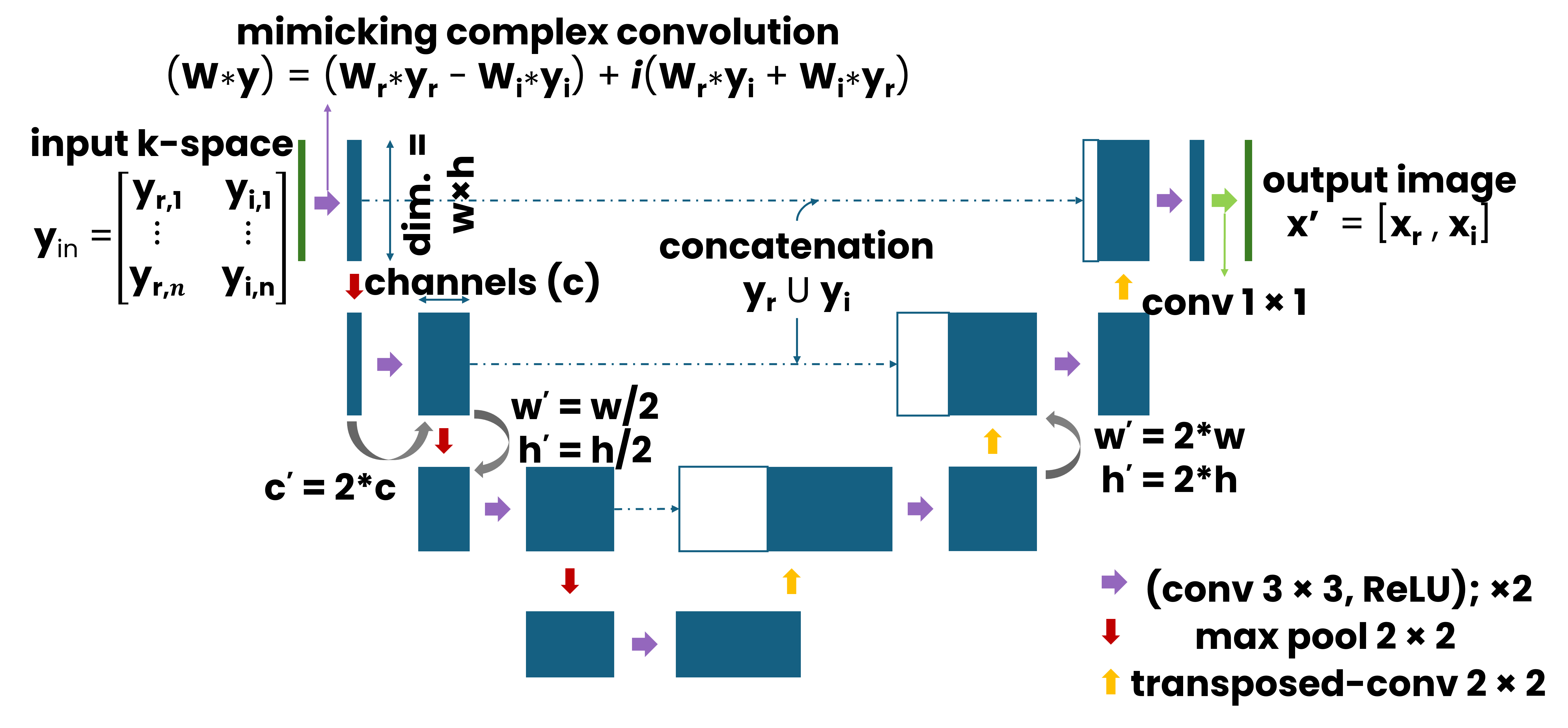}
\caption{K-space dual channel U-Net for LF-MRI SR/IQT using undersampled \emph{k}-space.} 
\label{fig:U-Net}
\vspace{-0.5cm}
\end{figure}

\vspace{-0.3cm}
To reconstruct HF-like \emph{k}-space from undersampled LF acquisitions, we employ a \emph{k}-space dual channel U-Net \cite{ronneberger2015u} that decomposes the input data into real $\mathcal{R}(\cdot)$ and imaginary $\mathcal{I}(\cdot)$ components, learning a two-channel nonlinear mapping, mathematically expressed as
\vspace{-0.1cm}
\begin{equation}
\hat{\mathbf{y}}_{\mathrm{HF}} = f_{\theta}\!\bigl(\mathcal{R}(\mathbf{y}_{\mathrm{LF}}^{\mathrm{us}}),\, \mathcal{I}(\mathbf{y}_{\mathrm{LF}}^{\mathrm{us}})\bigr),
\end{equation}
where $f_{\theta}$ denotes the hierarchical network with convolutional blocks and skip connections that preserve low-level structure, improve gradient flow, and maintain magnitude-phase consistency across layers. The schematic diagram of the proposed network is presented in Fig. \ref{fig:U-Net}, which shows an end-to-end workflow of our deep learning framework for SR/IQT from undersampled \emph{k}-space data. The backbone of our framework is a \emph{k}-space dual channel U-Net, specifically designed to process complex-valued data, represented as two real-valued channels by jointly handling the real and imaginary components. The network applies convolutional operations and activation functions to extract features from the \emph{k}-space dual channel, while interleaved pooling layers in the encoder enhance salient features and reduce spatial dimensions. The decoder leverages transposed convolutions and skip connections to progressively reconstruct high-resolution outputs while preserving the 2-channel structure of the \emph{k}-space input, ensuring that both magnitude and phase information are retained. The complex convolutions are approximated via algebraic combinations of real-valued operations
\begin{equation}
\quad (\mathbf{W} \ast \mathbf{y}) = (\mathbf{W_r} \ast \mathbf{y_r} - \mathbf{W_i} \ast \mathbf{y_i}) + i(\mathbf{W_r} \ast \mathbf{y_i} + \mathbf{W_i} \ast \mathbf{y_r}),
\end{equation}
where \(\mathbf{W_r}, \mathbf{W_i}\) are the real and imaginary parts of the convolutional weights, respectively, and \(\mathbf{y_r}, \mathbf{y_i}\) are the corresponding observations. This formulation ensures that the network preserves amplitude-phase relationships valuable for accurate \emph{k}-space reconstruction and high-fidelity image enhancement. 

For training the proposed architecture, the network learns to reconstruct HF-like \emph{k}-space from undersampled LF acquisitions using paired input-output examples generated by transforming HF images into synthetic LF counterparts with randomised, tissue-specific contrasts, capturing variability typical of real LF scans. The corresponding \emph{k}-space representations are then undersampled at varying acceleration rates using binary masks like pseudo-radial or Cartesian, simulating accelerated MRI acquisition. These paired undersampled LF and fully sampled HF \emph{k}-space datasets enable the network to jointly learn the mapping required for accurate frequency-domain reconstruction and enhancement of image quality. The trained model is then used to reconstruct HF-like MR image using the undersampled LF \emph{k}-space input.

\vspace{-0.3cm}
\subsection{Deep learning architecture configuration}
\label{ssec:arch}
\vspace{-0.25cm}
Figure \ref{fig:U-Net} shows a modified three-layer U-Net architecture \cite{ronneberger2015u} designed for complex-valued MRI \emph{k}-space reconstruction. The \emph{k}-space is represented as two real-valued channels and processed as floating-point tensors. The network jointly handles the real and imaginary components of the input slices, mapping them to a two-channel HF-like \emph{k}-space output. Standard convolutional layers are used to provide a consistent baseline for comparing models trained in both the spatial and \emph{k}-space domains. The encoder consists of three convolutional blocks: the first block contains two consecutive $3 \times 3$ convolutional layers with 64 output channels each, followed by a $2 \times 2$ max-pooling layer; the second block has two $3 \times 3$ convolutions with 128 output channels each, followed by max-pooling; and the third block has two $3 \times 3$ convolutions with 256 output channels each, again followed by max-pooling. The bottleneck consists of two $3 \times 3$ convolutional layers with 512 channels, capturing the most abstract representations. The decoder mirrors the encoder, with three upsampling stages: the first stage uses a $2 \times 2$ transposed convolution to reduce the channel depth from 512 to 256, followed by two consecutive $3 \times 3$ convolutional layers with 256 channels concatenated with the corresponding encoder block via skip connections; the second stage upsamples from 256 to 128 channels, followed by two consecutive $3 \times 3$ convolutional layers with 128 channels concatenated with the second encoder block; and the third stage upsamples from 128 to 64 channels, followed by two consecutive $3 \times 3$ convolutional layers with 64 channels concatenated with the first encoder block. All convolutional layers employ ReLU activations except the final $1 \times 1$ convolution, which outputs two channels for the reconstructed real and imaginary \emph{k}-space components.
\vspace{-0.5cm}
\section{Experiments and Results}
\vspace{-0.3cm}
\subsection{Datasets}
\label{ssec:data}
\vspace{-0.25cm}
We evaluate the performance of our proposed approach using high-resolution T1-weighted (T1w) volumes from the WU-Minn Human Connectome Project (HCP) \cite{Sotiropoulos2013Oct}, acquired at 3T with $0.7$\,mm isotropic voxels. We randomly select 40 subjects (corresponding to 4000 slices) for training/validation and hold out 10 subjects (corresponding to 200 slices) for testing. The LF counterparts are synthesised from the HF images using the stochastic LF simulator in \cite{Hongxiang2022}, which samples tissue-specific contrast changes from a learned prior to mimic LF SNR/contrast degradation. To emulate accelerated acquisition, we undersample the LF \emph{k}-space with pseudo-radial and Cartesian sampling patterns at sampling rates of 50\% and 30\%, and compare with their fully-sampled 100\% references. Reconstructions are then compared across the same acceleration levels in both spatial and \emph{k}-space IQT settings. 
\vspace{-0.3cm}
\subsection{Training and Testing}
\label{ssec:training}
\begin{figure}
\vspace{-0.3cm}
\centering
\includegraphics[width=0.995\columnwidth]{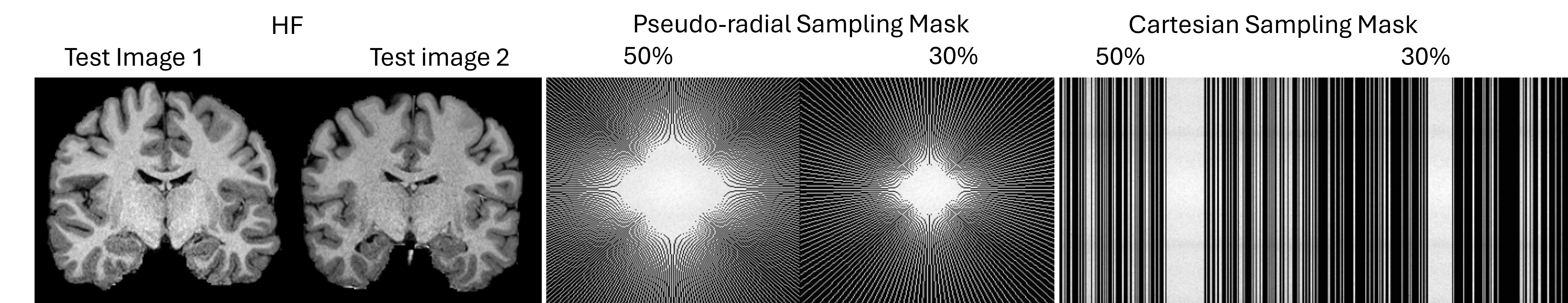}
\vspace{-0.8cm}
\caption{HF-MR test images, and pseudo-radial and cartesian undersampling binary masks at 50\% and 30\% sampling rates.} 
\label{fig:PRS_HFslices}
\vspace{-0.3cm}
\end{figure}

\vspace{-0.25cm}
The proposed network is trained using the Adam optimizer with a learning rate of $10^{-3}$ and a weight decay of $10^{-6}$, a batch size of 8, and a combined mean absolute error (MAE) and mean squared error (MSE) loss to balance robustness and reconstruction fidelity. To enhance generalisation, we employ a 3-fold cross-validation strategy on 4000 training slices, where in each fold one third of the slices is used for validation and the remaining two thirds are used for training. Training is conducted for 150 epochs per fold, and the best model checkpoint is selected based on the lowest validation loss. All experiments are implemented in PyTorch and executed on an NVIDIA A100 GPU with 80\,GB of memory. The performance of the proposed approach is evaluated using 100 held-out slices. During inference, each slice is reconstructed independently at each sampling rate, and reconstruction quality is quantified using peak signal-to-noise ratio (PSNR) and structural similarity index measure (SSIM) against HF references. Figure \ref{fig:PRS_HFslices} presents two HF test images along with their corresponding pseudo-radial and Cartesian sampling masks at 50\% and 30\% sampling rates, respectively.
\vspace{-0.3cm}
\subsection{Quantitative analysis}
\vspace{-0.25cm}
Table \ref{tab:PSRiqt} provides the PSNR and SSIM values for IQT at different undersampling ratios, using both pseudo-radial and Cartesian sampling patterns. These results are compared to those obtained from IQT using spatial domain images. As expected, the baseline interpolated images exhibit poor SSIM and lower PSNR across both domains, with performance degrading further as the sampling rate decreases. For example, in case of pseudo-radial sampling, SSIM drops from 0.8473 at full sampling to 0.3859 at 30\% sampling, reflecting the severe loss of structural information under heavy undersampling. On the other hand, both IQT approaches substantially improve over the LF baseline, confirming the effectiveness of IQT in recovering HF-like image features. Notably, even at 30\% sampling, spatial-domain IQT raises SSIM to 0.9330 compared to 0.3859 for the LF input, demonstrating the strong corrective power of the method. Between the two IQT strategies, and across all sampling rates, \emph{k}-space IQT consistently achieves higher SSIM and PSNR values compared to spatial IQT. While the improvements over spatial-domain IQT are small at full sampling, they become more evident under undersampling. In particular, at 30\% sampling, \emph{k}-space IQT achieves 0.9406 SSIM and 33.17 dB PSNR, outperforming the spatial-domain counterpart (0.9330 SSIM and 30.54 dB PSNR). This suggests that aggressive undersampling of \emph{k}-space can be leveraged to accelerate MRI acquisition times while still enabling high-quality reconstructions via IQT. Thus, this highlights the benefit of learning directly in \emph{k}-space rather than applying IQT only as a post-processing step after image reconstruction.

\begin{table}
\vspace{-0.3cm}
\caption{Average and standard deviation PSNR and SSIM measures for \emph{k}-space and spatial domain IQT at different sampling rates. Bold indicates best results.}
\centering
\footnotesize
\setlength{\tabcolsep}{3pt}
\renewcommand{\arraystretch}{1.1}
\resizebox{\linewidth}{!}{%
\begin{tabular}{l|l|c|c|c}
\hline
\textbf{Domain} & \textbf{Metric} & \textbf{100\%} & \textbf{50\%} & \textbf{30\%} \\
\hline\hline
\multicolumn{5}{c}{\textbf{Pseudo-radial Sampling}} \\
\hline\hline
\textbf{Interpolation} & SSIM & 0.8473 $\pm$ 0.0197 & 0.4946 $\pm$ 0.0303 & 0.3859 $\pm$ 0.0153 \\
 & PSNR & 25.41 $\pm$ 1.6792 & 21.64 $\pm$ 0.8683 & 21.42 $\pm$ 0.6350 \\
\hline\hline
\textbf{IQT Spatial} & SSIM & 0.9499 $\pm$ 0.0116 & 0.9422 $\pm$ 0.0116 & 0.9330 $\pm$ 0.0137 \\
 & PSNR & 33.85 $\pm$ 2.1470 & 31.45 $\pm$ 1.8278 & 30.54 $\pm$ 1.9291 \\
\hline\hline
\textbf{IQT K-Space} & SSIM & \textbf{0.9505 $\pm$ 0.0118} & \textbf{0.9480 $\pm$ 0.0122} & \textbf{0.9406 $\pm$ 0.0141} \\
 & PSNR & \textbf{33.88 $\pm$ 2.1624} & \textbf{33.70 $\pm$ 2.1303} & \textbf{33.17 $\pm$ 2.1178} \\
\hline\hline
\multicolumn{5}{c}{\textbf{Cartesian Sampling}} \\
\hline\hline
\textbf{Interpolation} & SSIM & 0.8473 $\pm$ 0.0197 & 0.4945 $\pm$ 0.0302 & 0.3858 $\pm$ 0.0152 \\
 & PSNR & 25.41 $\pm$ 1.6792 & 21.63 $\pm$ 0.8682 & 21.41 $\pm$ 0.6349 \\
\hline\hline
\textbf{IQT Spatial} & SSIM & 0.9499 $\pm$ 0.0116 & 0.9421 $\pm$ 0.0115 & 0.9329 $\pm$ 0.0136 \\
 & PSNR & 33.85 $\pm$ 2.1470 & 31.44 $\pm$ 1.8277 & 30.53 $\pm$ 1.9290 \\
\hline\hline
\textbf{IQT K-Space} & SSIM & \textbf{0.9505 $\pm$ 0.0118} & \textbf{0.9479 $\pm$ 0.0121} & \textbf{0.9405 $\pm$ 0.0140} \\
 & PSNR & \textbf{33.88 $\pm$ 2.1624} & \textbf{33.69 $\pm$ 2.1302} & \textbf{33.16 $\pm$ 2.1177} \\
\hline
\end{tabular}}
\label{tab:PSRiqt}
\vspace{-0.35cm}
\end{table}

\begin{figure}
\vspace{-0.3cm}
\centering
\includegraphics[width=0.995\columnwidth]{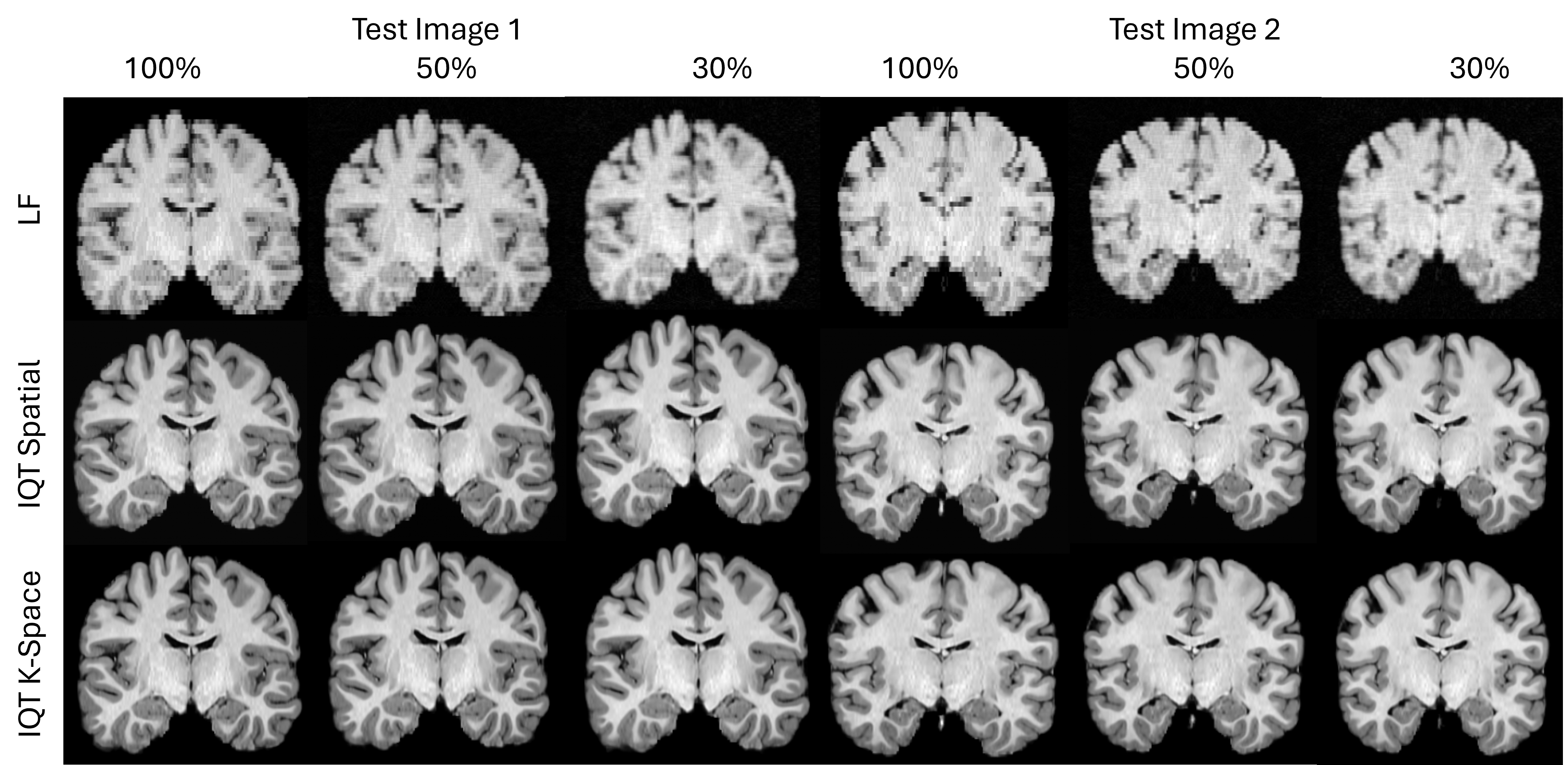}
\vspace{-0.6cm}
\caption{Reconstruction results of the LF images, and IQT in the spatial and \emph{k}-space domains, at different pseudo-radial under-sampling rates using two test images.} 
\label{fig:PRS_IQTslices}
\vspace{-0.3cm}
\end{figure}
\vspace{-0.3cm}
\begin{figure}
\vspace{-0.4cm}
\centering
\includegraphics[width=0.995\columnwidth]{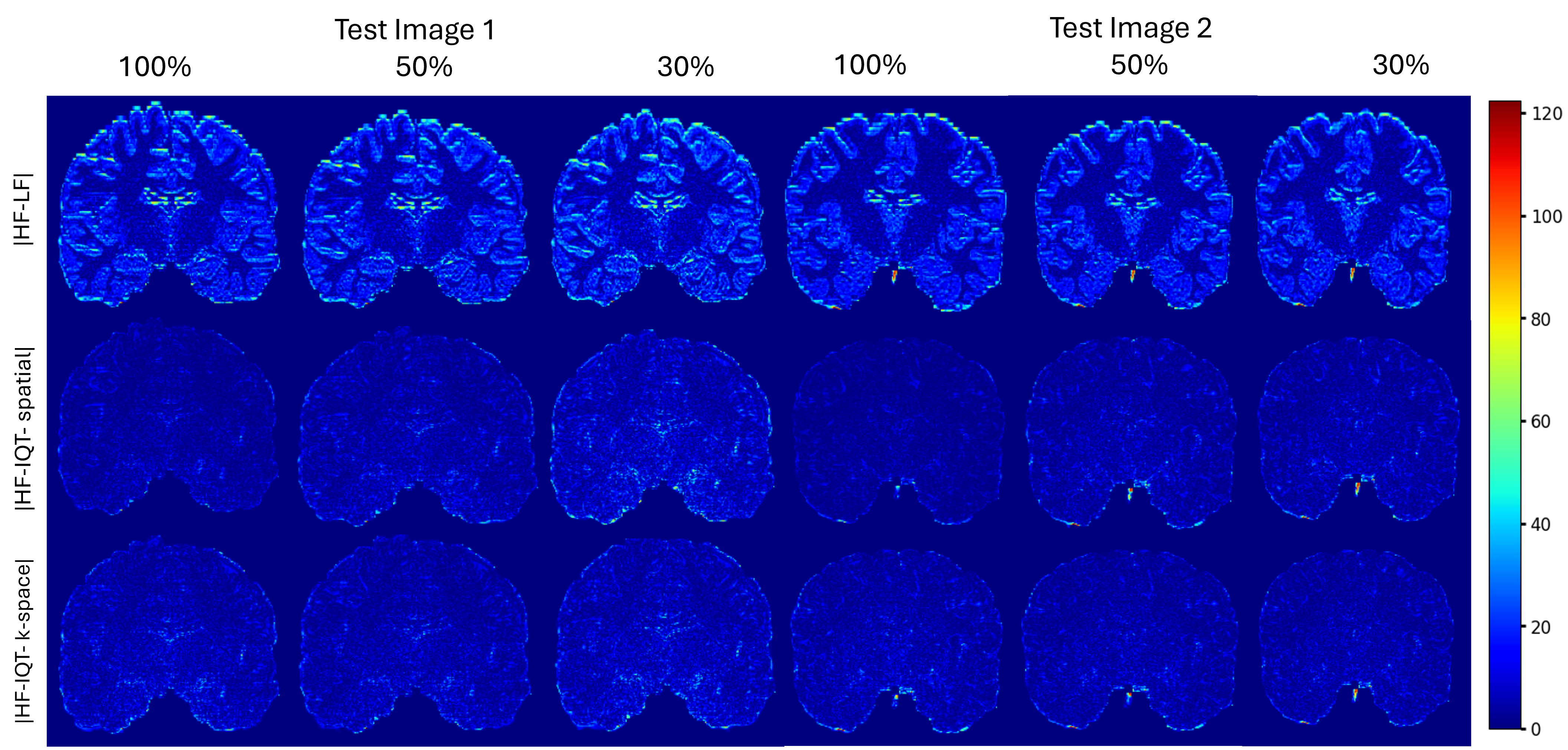}
\vspace{-0.6cm}
\caption{Absolute error maps between HF images in Fig. \ref{fig:PRS_HFslices}, and the reconstructions in Fig. \ref{fig:PRS_IQTslices} at different pseudo-radial under-sampling rates: $|\text{HF} - \text{LF}|$ in row 1, $|\text{HF} - \text{IQT-spatial}|$ in rows 2 and $|\text{HF} - \text{IQT-\emph{k}-space}|$ in row 3.} 
\label{fig:PRS_EMslices}
\end{figure}
\subsection{Qualitative analysis}
\vspace{-0.25cm}
To further assess the effectiveness of the proposed approach, Fig. \ref{fig:PRS_IQTslices} displays two reconstructed brain test images from the HCP dataset, generated using both \emph{k}-space and spatial domain approaches with pseudo-radial sampling at different undersampling rates. Additionally, Figure \ref{fig:PRS_EMslices} presents error maps which show the absolute differences between the original HF images in Fig. \ref{fig:PRS_HFslices} and their corresponding LF and IQT counterparts. The results obtained with Cartesian sampling exhibit similar trends and are therefore not included here. The qualitative results align with the quantitative findings in Table \ref{tab:PSRiqt}. As expected, interpolation reconstructions exhibit significant visual degradation at lower sampling rates, with noticeable blurring and loss of fine details. The degradation is particularly evident at 30\% sampling, where structural distortions and intensity inconsistencies become more pronounced. The corresponding error maps reveal relatively high residual errors, especially around anatomical edges, indicating substantial reconstruction inaccuracies. In contrast, IQT-based reconstructions consistently restore finer details, yielding images with enhanced structural fidelity and reduced visual artifacts. Even at 30\% sampling, IQT maintains a clear depiction of anatomical structures with minimal blurring. The error maps further validate this improvement, showing significantly lower residual errors compared to LF reconstructions. Comparing \emph{k}-space and spatial domain reconstructions, we observe a comparable performance at full \emph{k}-space IQT but the differences become more evident as the sampling rate increases (i.e., 50\% and 30\%). 
\vspace{-0.35cm}
\section{Conclusion}
\vspace{-0.35cm}
In this work, a novel \emph{k}-space-based super-resolution (SR)/image quality transfer (IQT) framework has been proposed to jointly reconstruct high-field-like magnetic resonance images from undersampled \emph{k}-space data, unlike conventional methods, which treat reconstruction and enhancement as separate tasks. By leveraging \emph{k}-space information rather than relying on spatial-domain post-processing, the proposed approach allows for \emph{k}-space under-sampling, which in turn accelerates scan times while maintaining high diagnostic quality. Experiments on low-field brain MRI dataset demonstrated that \emph{k}-space-driven SR/IQT outperforms its spatial-domain counterpart while also enabling significant scan time acceleration, as undersampled \emph{k}-space reconstructions achieved quality comparable to fully sampled acquisitions. Future work will explore adaptive \emph{k}-space sampling strategies guided by uncertainty maps to optimise acquisition efficiency and maintain high-quality reconstructions.

\vspace{-0.25cm}
\section{COMPLIANCE WITH ETHICAL STANDARDS}
\vspace{-0.25cm}
This research study was conducted retrospectively using a publicly available brain MRI scan dataset \cite{Sotiropoulos2013Oct}. Ethical approval was not required.

\vspace{-0.4cm}
\bibliographystyle{IEEEbib}
\bibliography{strings,refs}

\end{document}

%% file: com_notations.tex
\def\bfOmega{{\boldsymbol{\Omega}}}

\def\bs0{{\boldsymbol{0}}}

\def\bfx{{\mathbf{x}}}
\def\bfy{{\mathbf{y}}}

\def\bfF{{\mathbf{F}}}



%% file: notations_yoann.tex
\input com_notations.tex

\newcounter{algo}
\renewcommand{\thealgo}{\arabic{algo}}